\documentclass[runningheads]{llncs}

\usepackage{eccv}

\usepackage{eccvabbrv}
\usepackage{enumitem}
\usepackage{graphicx}
\usepackage{booktabs}

\usepackage[accsupp]{axessibility}  

\usepackage[linesnumbered,ruled,vlined]{algorithm2e}
\usepackage{xcolor}
\usepackage{multicol} 
\definecolor{darkred}{rgb}{0.6, 0.0, 0.0}  

\usepackage{amsmath} 

\usepackage{titlesec}

\usepackage[pagebackref,breaklinks,colorlinks,citecolor=eccvblue]{hyperref}

\usepackage{orcidlink}

\usepackage{hyperref} 

\begin{document}

\title{PackMamba: Efficient Processing of Variable-Length Sequences in Mamba Training} 

\titlerunning{PackMamba: Efficient Var-Length Mamba Training}

\author{
    Haoran Xu\inst{1} \and
    Ziqian Liu\inst{1,2} \and
    Rong Fu\inst{1} \and
    Zhongling Su\inst{1} \and
    Zerui Wang\inst{1} \and
    Zheng Cai\inst{1} \and
    Zhilin Pei\inst{1} \and
    Xingcheng Zhang\inst{1}
}
\authorrunning{H. Xu, Z. Liu, R. Fu et al.}


\institute{
    Shanghai Artificial Intelligence Laboratory, China\\
    \email{\{xuhaoran, furong, suzhongling, wangzerui, caizheng, peizhilin, zhangxingcheng\}@pjlab.org.cn} \and
    Renmin University of China, China\\
    \email{liuziqian@ruc.edu.cn}
}

\maketitle
\vspace{-0.7cm}
\begin{center}
\small 
\href{https://github.com/ptxu78/pack_mamba}{https://github.com/ptxu78/pack\_mamba}
\end{center}
\vspace{-0.5cm}
\begin{abstract}
With the evolution of large language models, traditional Transformer models become computationally demanding for lengthy sequences due to the quadratic growth in computation with respect to the sequence length. 
Mamba, emerging as a groundbreaking architecture in the field of generative AI, demonstrates remarkable proficiency in handling elongated sequences with reduced computational and memory complexity. 
Nevertheless, the existing training framework of Mamba presents inefficiency with variable-length sequence inputs. Either single-sequence training results in low GPU utilization, or batched processing of variable-length sequences to a maximum length incurs considerable memory and computational overhead.
To address this problem, we analyze the performance of bottleneck operators in Mamba under diverse tensor shapes and propose PackMamba, a high-throughput Mamba that efficiently handles variable-length sequences. 
Diving deep into state-space models (SSMs), we modify the parallel operators to avoid passing information between individual sequences while maintaining high performance. 
Leveraging hardware-software co-optimization, this modification ensures coalesced memory access to position indices without extra kernel overhead.
Experimental results on an NVIDIA A100 GPU demonstrate throughput exceeding the baseline single-sequence processing scheme: 3.06x speedup on the 1.4B model and 2.62x on the 2.8B model.
\end{abstract}

\vspace{-0.1cm}
\section{Introduction}
\label{section:1}


\indent With the swift advancement of large-scale language models, the Transformer, serving as the most widely used foundational model in large models, has been applied in BERT~\cite{bert}, GPT-4~\cite{gpt4}, and Llama~\cite{llama}. The effectiveness of self-attention stems from its capacity to facilitate dense information exchange within a given context window, thereby enabling it to model intricate data. Nevertheless, this characteristic introduces inherent limitations, namely the incapacity to model elements outside of a bounded window~\cite{mamba-main}, and the computational complexity that scales quadratically with the increase in window length. To address these drawbacks inherent to the Transformer model, an increasing amount of research is being directed towards the innovation of foundational models in natural language processing (NLP), such as linear attention~\cite{linear_attention}, gated convolution~\cite{gated_conv}, recurrent models~\cite{recurrent},  structured state space models~\cite{State_space_models}(SSMs) and Test-Time Training~\cite{TTT}.
Mamba~\cite{mamba-main} enjoys fast inference (5× higher throughput than Transformers) and linear scaling in sequence length, and its performance improves on real data up to million-length sequences. On language modeling, Mamba-3B model outperforms Transformers of the same size and matches Transformers twice its size, both in pretraining and downstream evaluation. 

In contrast to Recurrent Neural Networks (RNNs), which inherently rely on the state of preceding neurons, thereby impeding parallel training, Mamba innovatively introduces a 'selective scan' mechanism that promotes parallelism during the training process. This departure from the conventional sequential dependency in RNNs empowers Mamba to harness parallel computational resources, consequently boosting training efficiency and scalability.


Mamba's training faces challenges with variable-length sequence inputs. Profiling shows that in single-sequence training, GPU tasks are fine-grained, with large gaps between tasks. These gaps, caused by frequent CPU-GPU synchronization and high kernel launch overhead, lead to inefficient GPU utilization. Another approach is pad to maximum length for batch training. However, these padded zeros introduce significant computation and memory overhead.

\begin{figure}[h]
    \centering
    \includegraphics[width=\textwidth]{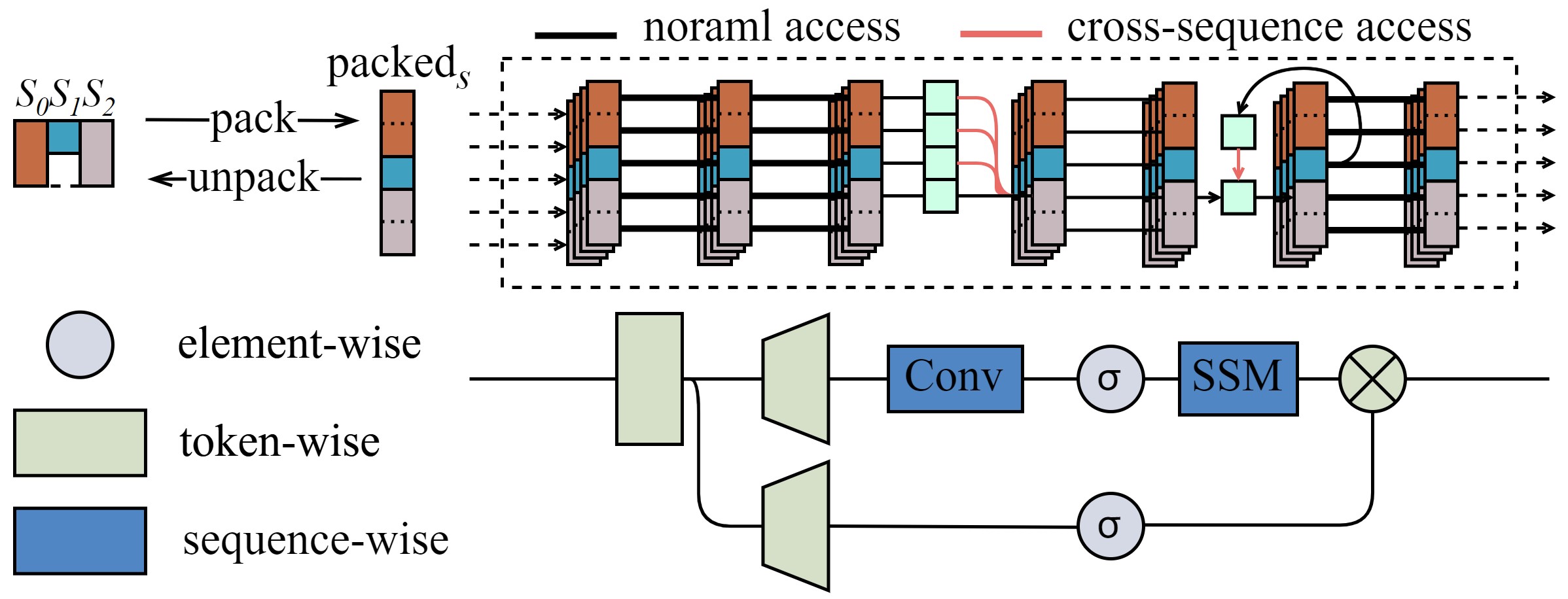}
    \caption{PackMamba overview}
    \label{fig:overview}
\end{figure}

To address the issue of low training throughput in Mamba, we introduce PackMamba - a high-performance Mamba that supports variable-length sequences. As shown in Figure~\ref{fig:overview}, PackMamba initially packs variable-length sequences into longer sequences (4096 in Mamba-1.3B) using indices to record the original sequence start positions within the long sequence. By modifying the selective scan and convolution operators to prevent the cross-sequence access (marked in red line), the training process is adapted to support variable-length sequences. Additionally, we employ techniques such as shared memory and memory access merging for a synergistic acceleration of GPU memory access, optimizing performance. These advancements ensure that PackMamba not only overcomes the limitations of Mamba but also significantly enhances its efficiency and scalability, thereby making it a more robust solution for neural network training.

To sum up, this paper makes the following contributions:
\begin{enumerate}
    \item We conduct an operator analysis of the SSM, the bottleneck of Mamba, and discover that hardware and the model demonstrate varying affinities towards sequence length.
    \item We propose a Mamba training method that accommodates variable-length sequences, realizing the support of convolutional layers and State Space Model layers for variable-length sequences.
    \item We conduct experiments on an NVIDIA A100 GPU, demonstrating a 3.06x speedup in throughput for the 1.4B model and a 2.62x speedup for the 2.8B model.
\end{enumerate}

\vspace{-0.1cm}
\section{Motivation}
\subsection{Variable-Length Inputs Train}
\label{section:2.1}
As depicted in Section~\ref{section:1}, a straightforward solution to solve fine-grained computation in variable-length training is to pad all sequences with zeros to the maximum sequence length, but this method leads to a significant waste of memory and computational resources. Our experiments demonstrate that the use of this padding method with the InternLM dataset~\cite{internlm1,internlm2} results in 66.3\% padding rate.

There has been some work focusing on the training issue of variable-length sequences in Transformers. Both Tencent~\cite{variable-length1} and Baidu~\cite{variable-length2} initially attempted to minimize the proportion of zero-padding by bundling sequences of similar lengths. ByteTransformer~\cite{bytetransformer} proposed a padding-free algorithm that packs the input tensor with variable-length sequences and calculates the positioning offset vector for all transformer operations to index. This approach, combined with software and hardware co-optimization, achieved good results. In addition, ByteTransformer also employs methods such as kernel fusion and CUTLASS grouped GEMM optimizations to enhance overall performance. 

However, this issue has not been resolved in Mamba. Unlike the Transformer, where the positioning offset can be used to obtain the actual training data of each sub-segment after unpacking, the selective scan operator in Mamba has a dependency where the latter state depends on the former one. This increases the difficulty of maintaining parallelism.
\subsection{SSM Operator Analysis}
\label{section:2.2}
\begin{figure}[h]
    \centering
    \includegraphics[width=\textwidth]{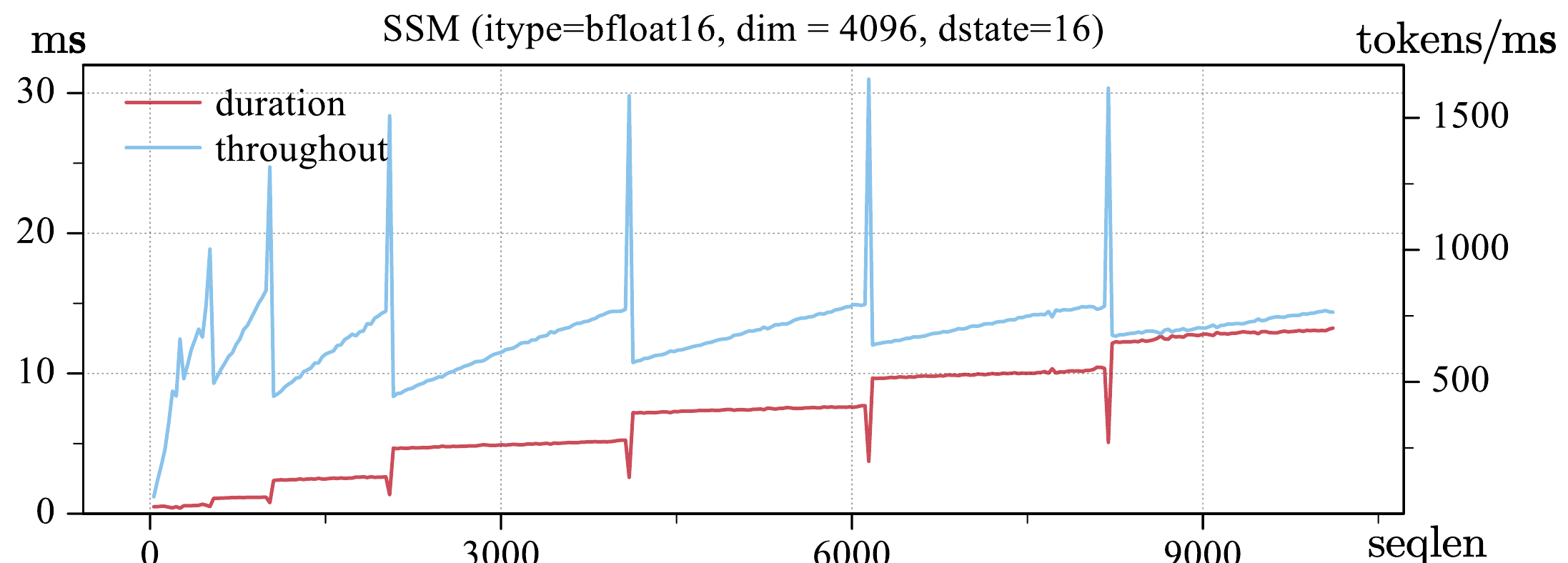}
    \caption{SSM profiling}
    \label{fig:SSM profiling}
\end{figure}

\begin{figure}[h]
    \centering
    \includegraphics[width=\textwidth]{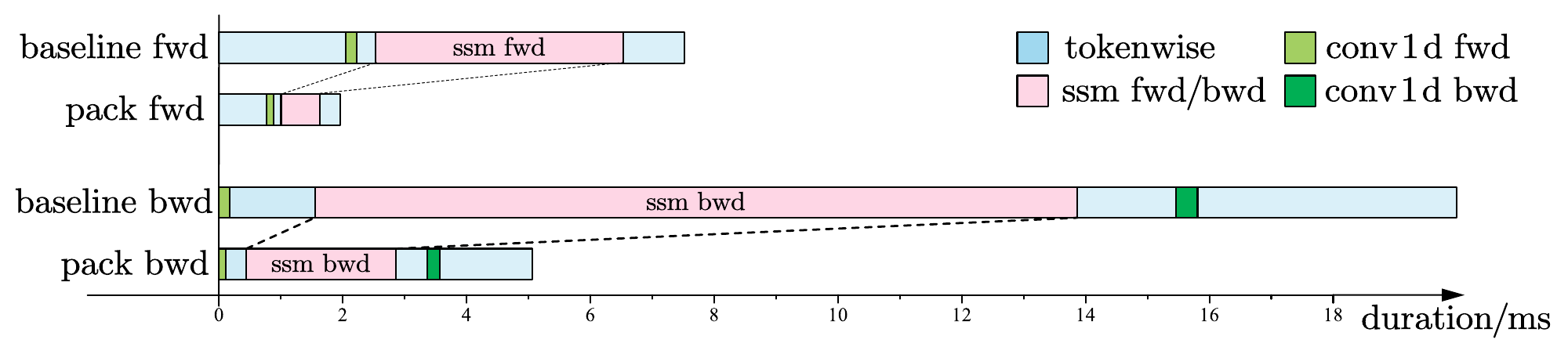}
    \caption{Kernel speedup(Mamba-1.4B, seqlen=4096)}
    \label{fig:time}
\end{figure}

As shown in Figure~\ref{fig:time}, in the padding approach, the SSM operator is a bottleneck, using 59.3\% of the process resources. Zero-padding along the sequence dimension wastes its computational resources. Therefore, we analyze the performance of the operator with different seqlen inputs. Analyzing Figure~\ref{fig:SSM profiling} and the details of the operator, we draw the following conclusions: 

\begin{enumerate}
    \item When $2^n < \text{seqlen} < 2^{n+1}$, the duration increases slowly with increasing \text{seqlen}. This is due to the operator's internal padding zeros.
    \item When $\text{seqlen} = 2^n$ (or a multiple of 2048), the processing time significantly decreases. This reduction is due to an internal vector loading mechanism which provides a speedup of 1.51x to 2.03x when activated.

    \item When $\text{seqlen} = 2^n$, as $n$ increases, the throughput logarithmically increases.
\end{enumerate}

Therefore, constructing input sequences such that $\text{seqlen} = 2^n$ can achieve high throughput. As for the external padding of zeros, one method is to pack multiple sequences into one long set, but passing states between sequences can lead to contamination of input data. We have analyzed and modified Mamba to ensure consistent training results before and after packing, thereby accelerating training.

\vspace{-0.5cm}


\vspace{-0.8cm}
\section{Main Approach}
\subsection{Packing-Unpacking Invariance}
\label{section:3.1}

\begin{figure}[h]
    \centering
    \includegraphics[width=\textwidth]{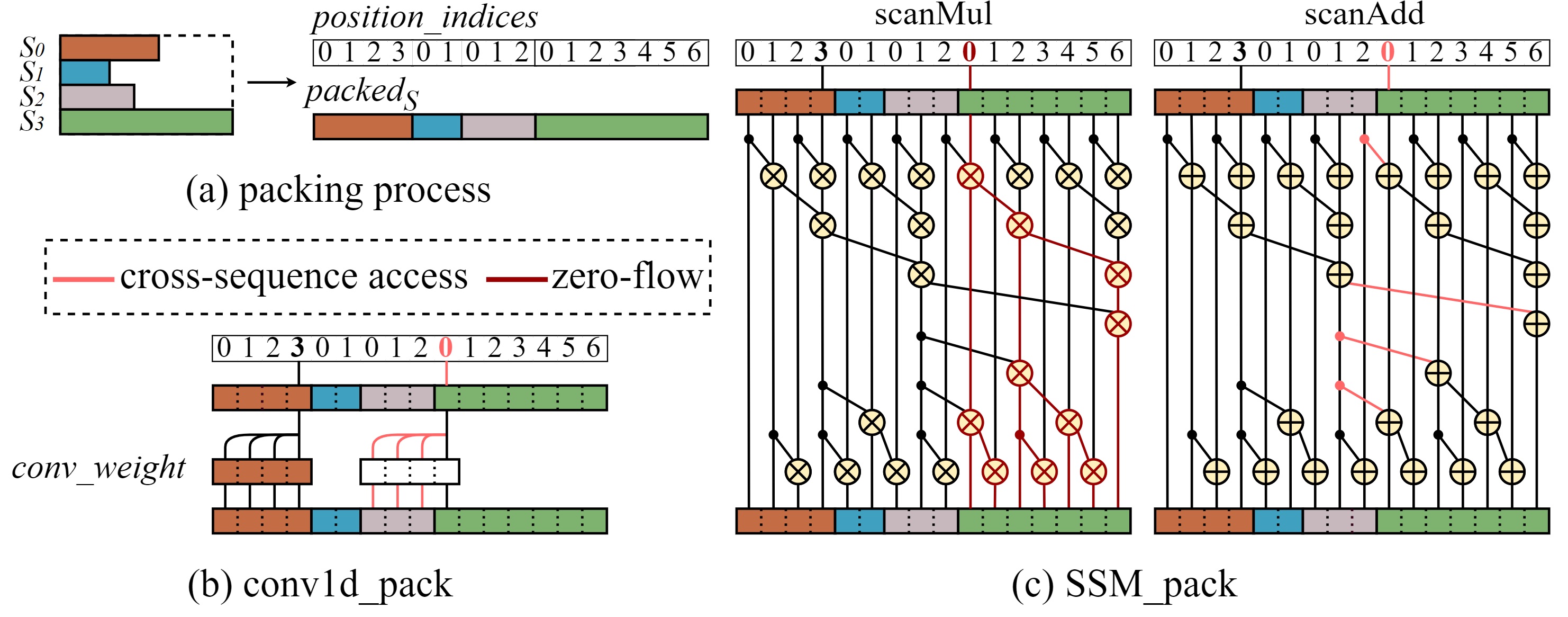}
    \caption{Mamba sequence-wise operators}
    \label{fig:approach}
\end{figure}

The \texttt{pack()} operation refers to concatenating the input tensor along the sequence dimension to obtain a \texttt{packed\_sequence} and the auxiliary structure \texttt{position\_indices} (detailed in Section~\ref{section:3.3}) as shown in Figure~\ref{fig:approach}(a). The \texttt{unpack()} operation is the inverse of \texttt{pack()}. 

A function \( f \) satisfies the \textbf{\emph{Packing-Unpacking Invariance}(PUI)} property if and only if for any data structure \( A \):
\[
f(S) = \texttt{unpack}(f(\texttt{pack}(S)))
\]

The property of PUI is transitive. If each sub-operation \( f_1, f_2, \ldots, f_n \) of a composite operation \( F \) satisfies this property, then \( F \) itself also does.

\subsection{Functional Analysis on Mamba Operators}
\label{section:3.2}
As shown in Figure~\ref{fig:overview}, the operations in the Mamba block can be categorized into three types:
\begin{enumerate}[label=(\roman*)]
    \item The sigmoid function is element-wise. Each scalar in the input tensor independently determines the output.
    \item Gemm (linear) and MSENorm are token-wise, not influencing each other across the sequence length dimension.
    \item Conv1d and SSM are sequence-wise, where adjacent tokens from the same sequence influence each other.
\end{enumerate}

The \texttt{pack()} and \texttt{unpack()} operations target the sequence length dimension and do not affect (i) element-wise and (ii) token-wise operators. Therefore, these two types of operators satisfy PUI. Sequence-wise operators do not satisfy PUI. In conv1d, the convolution kernel sliding across sequence boundaries leads to extra convolution items in subsequent sequences. In SSM, the state is also passed forward across the boundaries of sequences, leading to errors. Therefore, to ensure the Mamba blocks and the entire network satisfy PUI, modifications to the sequence-wise operations are necessary.
\begin{multicols}{2}
\noindent  
\scriptsize
\vspace{-0.5cm}
\begin{algorithm}[H]
\label{conv1d}
\caption{conv1d\_pack}
\KwIn{$x: (B, D, kNElts)$ \\ \hspace*{30pt} $indices: (B, kNElts)$ }
\KwOut{$y: (B, D, kNElts)$}
$width: (D, width) \leftarrow \text{Parameter}$ \\
\For{$i \leftarrow 0$ \KwTo $kNElts-1$}{
    \textcolor{darkred}{%
        \If{$indices[i] < width$}{
            $w \gets width - indices[i] - 1$\;
        }
    }
    \For{$j \leftarrow w$ \KwTo $width$}{
        $y[i] \gets y[i] + convTerm_j $
    }
}
\Return{$y$}

\end{algorithm}

\begin{algorithm}[H]
\caption{ScanOp\_pack}
\KwIn{$x: (B, D, L)$  $indices: (B, L)$\\ \hspace*{30pt} $\overline{A}, \overline{B}: (B, D, L)$}
\KwOut{$y: (B, D, L)$}
$\overline{B}: (B, D, L)\leftarrow \overline{B} \circ x$ \\
\textcolor{darkred}{%
    Set $\overline{A}[i]$ to zero when $indices[i]$ is zero. \\
}
\For{$step \leftarrow 0$ \KwTo $2\log_2(L) - 1$}{
    \text{scanAdd}($\overline{A}_{right}\circ \overline{B}_{left}, \overline{B}_{right}$)\\
    \text{scanMul}($\overline{A}_{left}, \overline{A}_{right}$)\\
}
\Return{$y$}
\end{algorithm}
\label{ssm}
\end{multicols}

\vspace{-0.7cm}

\subsection{Conv1d\_pack Implementation}
\label{section:3.3}
When convolving elements at the edges of a sequence, the convolution accesses the previous sequence, as illustrated by the red line in Figure~\ref{fig:approach}(b) for the first term of $S_3$. This cross-sequence access can be eliminated using \texttt{position\_indices}, which are generated during the \texttt{pack()} process and provide a way to track the original ordered positions of elements.

Algorithm~\ref{conv1d} outlines the modification for the forward process. As shown in red, the convolution for boundary elements (\texttt{index < width}) is terminated early. For the backward process, additional modifications are specifically necessary to calculate \(dx\) and \(dweight\). These modifications require reverse indices, which can be obtained from the position indices of the last \texttt{conv\_width} elements (detailed in Section~\ref{section:3.5}).

\vspace{-0.3cm}
\subsection{SSM\_pack Implementation}
\label{section:3.4}
In the SSM operator, state passing occurs at sequence boundaries in (1a), while selective SSMs can reset their state by setting $\Delta \to \infty$ in (2a). In serial mode, a simpler method is to directly set $\overline{A} \to 0$.
\begin{align}
    h_t &= \overline{A} h_{t-1} + \overline{B} x_t \hspace{10pt}(1a) && \quad \overline{A} = \exp(\Delta A) \hspace{85pt}(2a) \nonumber \\
    y_t &= C h_t \hspace{50pt}(1b) && \quad \overline{B} = (\Delta A)^{-1}(\exp(\Delta A) - I) \Delta B \hspace{9pt}(2b) \nonumber
\end{align}

The recurrence relation (1a) does not satisfy the associative property, but its expansion (3) can be realized stepwise through multiplication and addition, which are associative operations. Therefore, the parallel computation of SSM (Algorithm 2) is facilitated using two scan operators, as shown in Figure~\ref{fig:approach}(c).

\begin{equation}
h_t = \sum_{k=0}^t \left(\prod_{i=k+1}^t \overline{A}_i\right) x_k \tag{3} \nonumber
\end{equation}

Modifying the input of \texttt{scanMul} to set $\overline{A}_{position\_indices = 0} \to 0$ ensures this forward parallel operation with PUI too. This is because, if the $a^{th}$ term is set to zero, for the $m^{th}$ item processed by a single thread in \texttt{scanMul}: before the step, if the multiplied term $n > a$, then $\overline{A}[m] \neq 0$; if $n < a$, then $\overline{A}[m] = 0$. 

This change in \texttt{scanMul} affects every step of \texttt{scanAdd} by utilizing \(\overline{A}_{\text{right}} \circ \overline{B}_{\text{left}}\) on the input.
 For the $m^{th}$ item handled by a single thread in \texttt{scanAdd} (where $m > a$), if $n \geq a$, the current addend value remains non-zero and the accumulation proceeds normally; if $n < a$, the value is zero, thus preventing the state from previous terms before the $a^{th}$ item from passing through.

Similarly, the backward process consists of another two scan operators, where modifications only require setting $\overline{A}_{position\_indices = 0} \to 0$

\subsection{Software-hardware Co-optimization}
\label{section:3.5}
\begin{figure}[h]
    \centering
    \includegraphics[width=0.7\textwidth]{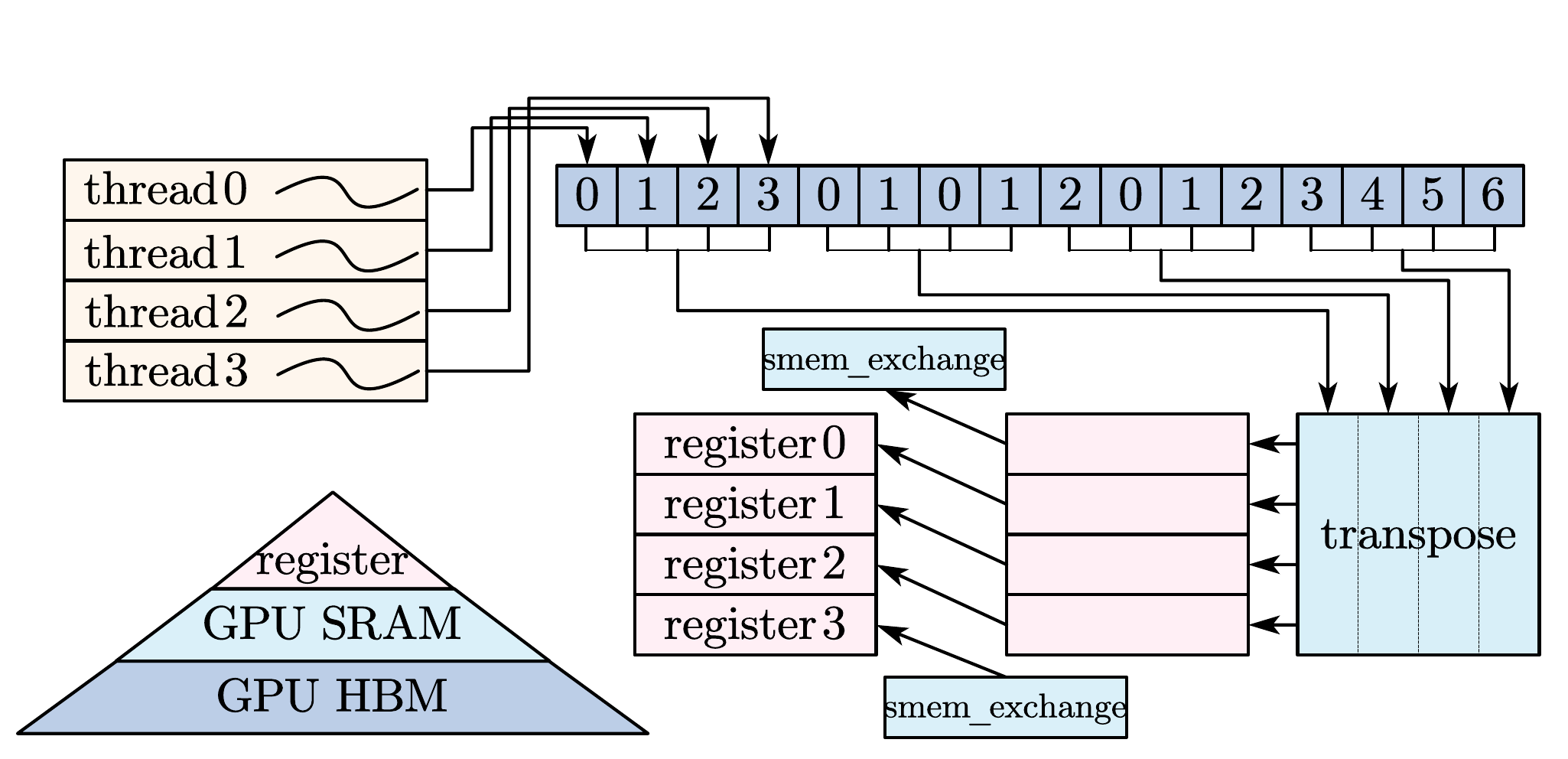}
    \caption{Memory Access Optimization}
    \label{fig:store}
\end{figure}
The modifying introduced additional overhead for reading and writing \\ \texttt{position\_indices}. We optimize the memory access logic and reused Mamba's structure for handling \texttt{hidden\_state} to significantly reduce additional read-write overhead and memory allocation costs.

As shown in Figure~\ref{fig:store}, during the reading process, continuous threads read the consecutive 
 \texttt{position\_indices} from HBM into SRAM. Within SRAM, the warp-striped arrangement is transposed into a blocked arrangement and transferred to the corresponding thread's registers, thereby participating in the computations. In the case of \texttt{conv\_bwd}, an additional step is required to stagger the data using SRAM to transfer reverse indices. This process incurs a cost of $n$ HBM reads (related to \texttt{seqlen}, up to 16 times), $2n$ SRAM read/write operations, and $n$ register writes. The main cost, reading from HBM, has been optimized through memory access coalescing. 

During computations, the use of \texttt{position\_indices} involves only register reads, which are negligible in terms of overhead.


\section{Evaluation}
We evaluate our approach on NVIDIA A100 GPUs. 
The training data, sourced from authentic corpora, consisted of sequences ranging in length from 57 to 2048, with an average length of 646. We utilize \texttt{bfloat16} and \texttt{float32} tensor types for inputs. The models trained were Mamba-110m with 16 layers and 1024 dimensions, Mamba-1.4B with 48 layers and 2048 dimensions, and Mamba-2.8B with 64 layers and 2560 dimensions. Training was executed using an 8-GPU data parallel. We compare three approaches: single-sequence, padding, and pack.

\begin{figure}[h]
    \centering
    \includegraphics[width=\textwidth]{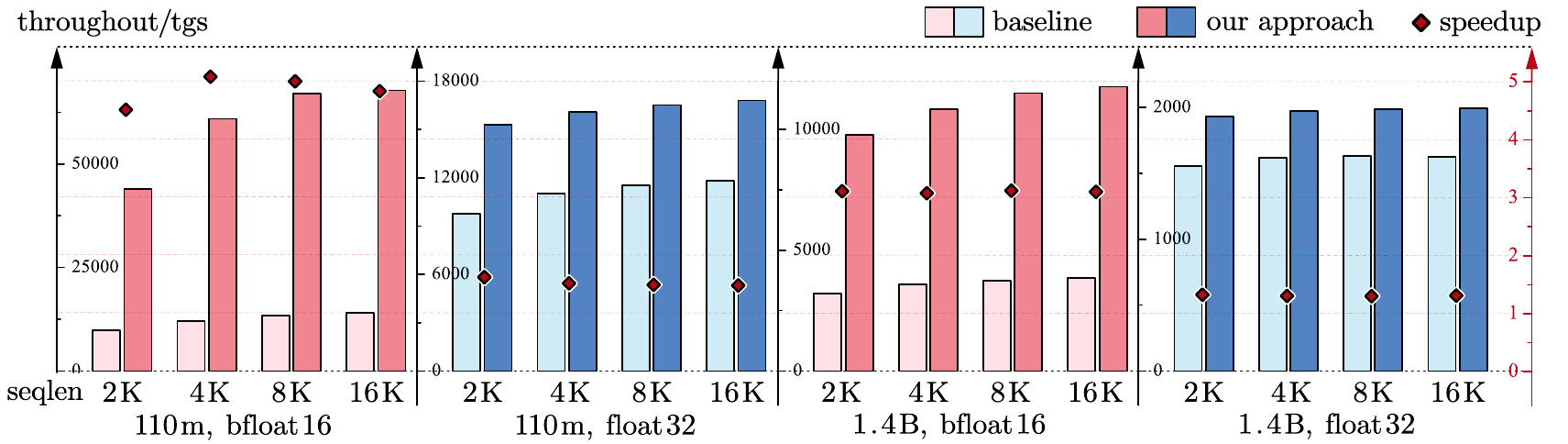}
    \caption{Training Throughput Comparison}
    \label{fig:throughput}
\end{figure}

\textbf{Throughput comparison.}\quad In training, we compute the average throughput of a stable sequence of 100 consecutive steps as the result. The baseline was set as the single-sequence approach, which consistently outperforms padding approach in throughput under all conditions. As shown in Figure~\ref{fig:throughput}, our approach accelerates performance by 3.06x to 5.05x compared to the baseline when using \texttt{itype=bfloat16} and by 1.34x to 1.57x with \texttt{itype=float32}. When scaling the model up to 2.8B, there is still a 2.61x speedup, demonstrating our approach's excellent scalability with larger models. 

\textbf{Kernel Speedup Analysis.}\quad We switch the baseline to the padding approach, as the kernel duration of the single-sequence approach is too sparse for reliable comparison. As shown in Figure~\ref{fig:throughput}, the forward-backward process achieves a 3.91x speedup, primarily due to the reduction in GEMM and SSM kernel durations. The increased density of the packed sequences eliminates significant amounts of idle computations.

\vspace{-0.5cm}

\vspace{-1cm}
\section{Discussion}

As described in Section~\ref{section:3.1}, the packing method involves sequentially packing sequences in the received order, sealing the pack when it cannot fit the next sequence. This method results in an average padding rate of 19.1\% on the InternLM dataset. By using a local greedy algorithm that sorts some of the sequences before packing, the padding rate can be reduced to as low as 0.41\%. However, this method incurs additional sorting time overhead.

PackMamba supports packing training with variable-length sequences while ensuring mathematical equivalence. Moreover, there are no instances of sequences spanning across packed sequences. In future work, we plan to address this issue by allowing long sequences to be cut at the end into two parts while still passing states between these parts. This approach will reduce padding to zero while maintaining data integrity and even support parallel strategies for infinitely long sequences.


%
%
\bibliographystyle{splncs04}
\bibliography{main}
\end{document}